\documentclass[10pt,journal,compsoc]{IEEEtran}

\usepackage{fix-cm}

\ifCLASSOPTIONcompsoc
  \usepackage[nocompress]{cite}
\else
  \usepackage{cite}
\fi

\ifCLASSINFOpdf
\else
\fi

\ifCLASSOPTIONcompsoc
 \usepackage[caption=false,font=footnotesize,labelfont=sf,textfont=sf]{subfig}
\else
 \usepackage[caption=false,font=footnotesize]{subfig}
\fi

\usepackage{amsmath,amsfonts}
\usepackage{algorithmic}
\usepackage{algorithm}
\usepackage{array}
\usepackage{textcomp}
\usepackage{marvosym}
\usepackage{stfloats}
\usepackage{url}
\usepackage{verbatim}
\usepackage{graphicx}
\graphicspath{ {figures/} }
\usepackage{cite}

\usepackage{booktabs}
\usepackage{color}
\usepackage[normalem]{ulem}
\usepackage{multirow}
\usepackage{bm}

\usepackage[pagebackref,breaklinks,colorlinks]{hyperref}

\usepackage{xspace}
\makeatletter
\newcommand*{\rom}[1]{\expandafter\@slowromancap\romannumeral #1@}

\DeclareRobustCommand\onedot{\futurelet\@let@token\@onedot}
\def\@onedot{\ifx\@let@token.\else.\null\fi\xspace}

\def\eg{\emph{e.g}\onedot} 
\def\ie{\emph{i.e}\onedot} 
 
 \def\vs{\emph{vs}\onedot}

\makeatother

\usepackage{color}

\newcommand{\minhl}[1]{#1}
\newcommand{\red}[1]{#1}
\newcommand\rurl[1]{\href{https://#1}{\nolinkurl{#1}}}
\newcommand\rdoi[1]{\href{https://doi.org/#1}{\nolinkurl{#1}}}

\begin{document}
\title{Explicit Correspondence Matching for Generalizable Neural Radiance Fields}

\author{Yuedong Chen\textsuperscript{\Letter}, Haofei Xu, Qianyi Wu, Chuanxia Zheng, Tat-Jen Cham, Jianfei Cai~\IEEEmembership{Fellow,~IEEE}
\thanks{Y. Chen (yuedong.chen@monash.edu), Q. Wu and J. Cai are with Monash University, Australia. H. Xu is with ETH Zurich, Switzerland. C. Zheng is with University of Oxford, UK. T.-J. Cham is with Nanyang Technological University, Singapore. DOI: \rdoi{10.1109/TPAMI.2025.3598711}}%
}

\markboth{~}%
{Chen \MakeLowercase{\textit{et al.}}: Explicit Correspondence Matching for Generalizable Neural Radiance Fields}

\IEEEtitleabstractindextext{%
\begin{abstract}
We present a new generalizable NeRF method that is able to directly generalize to new unseen scenarios and perform novel view synthesis with as few as two source views.
The key to our approach lies in the explicitly modeled correspondence matching information, so as to provide the geometry prior to the prediction of NeRF color and density for volume rendering. The explicit correspondence matching is quantified with the cosine similarity between image features sampled at the 2D projections of a 3D point on different views, which is able to provide reliable cues about the surface geometry.
Unlike previous methods where image features are extracted independently for each view, we consider modeling the cross-view interactions via Transformer cross-attention, which greatly improves the feature matching quality.
Our method achieves state-of-the-art results on different evaluation settings, with the experiments showing a strong correlation between our learned cosine feature similarity and volume density, demonstrating the effectiveness and superiority of our proposed method. The code and model are on our project page: \rurl{donydchen.github.io/matchnerf}.
\end{abstract}

}

\maketitle

\IEEEdisplaynontitleabstractindextext

\IEEEpeerreviewmaketitle

\IEEEraisesectionheading{\section{Introduction}\label{sec:intro}}

\IEEEPARstart{I}{n} the past few years, we have seen rapid advances in photorealistic novel view synthesis with Neural Radiance Fields (NeRF)~\cite{mildenhall2020nerf}\minhl{, Light Field Network (LFN)~\cite{sitzmann2021light} and 3D Gaussian Splatting (3DGS)~\cite{kerbl20233d}}.
However, \minhl{these vanilla~\cite{mildenhall2020nerf,sitzmann2021light,kerbl20233d} representations and their variants~\cite{barron2021mip,barron2022mip,mildenhall2022nerf,verbin2022ref,yu2024mip,huang20242d}} are mainly designed for per-scene optimization scenarios, where the lengthy optimization time and the need for vast amounts of views for each scene limit their practical usages in real-world applications. 

We are interested in a \emph{generalizable} NeRF, which aims at learning to model the generic scene structure \minhl{in a feed-forward manner} by conditioning the NeRF inputs on additional information derived from images~\cite{wang2021ibrnet,chibane2021stereo,chen2021mvsnerf}.
This form has a wide range of applications, because it can generalize to new unseen scenarios, and can perform reasonably well with just a few (\eg, 3) camera views, while it \emph{does not require any retraining}.

The existing generalizable NeRF approaches~\cite{yu2021pixelnerf,wang2021ibrnet,trevithick2021grf,chibane2021stereo,chen2021mvsnerf,johari2022geonerf,liu2022neural} generally adopt the pipeline of an image feature encoder that embeds multi-view images into a latent $\bm z$, and a NeRF decoder that conditions on $\bm z$ to predict the 3D radiance field and volume-renders it to generate the target-view image.
The key difference of these methods mainly lies in how the additional latent geometry is encoded.
Pioneer generalizable NeRF approaches~\cite{yu2021pixelnerf,wang2021ibrnet,trevithick2021grf} directly use and/or aggregate 2D convolutional image features independently extracted from each input source view, which struggle in new unseen scenarios, because their convolutional features are \emph{not explicitly geometry-aware}. 
To address this issue, recent works~\cite{chibane2021stereo,chen2021mvsnerf,johari2022geonerf,liu2022neural,xu2022point} incorporate the geometry-aware multi-view consistency to encode the geometry prior.
Among them, MVSNeRF~\cite{chen2021mvsnerf} is the most representative one, which 
constructs a plane-sweep 3D cost volume followed by a 3D Convolutional Neural Network (CNN) to generate the geometry prior $\bm z$, leveraging the success in multi-view stereo architectures~\cite{yao2018mvsnet,gu2020cascade}.
However, the construction of cost volume in MVSNeRF \emph{relies on a predefined reference view}, resulting in poor performance (see Fig.~\ref{fig:refviews}) if the target view does not have sufficient overlap with the reference view.
MVSNeRF also suffers from noisy backgrounds (see Fig.~\ref{fig:comp}) due to the cost volume.

\begin{figure}[t!]
    \begin{center}
    \includegraphics[width=\linewidth]{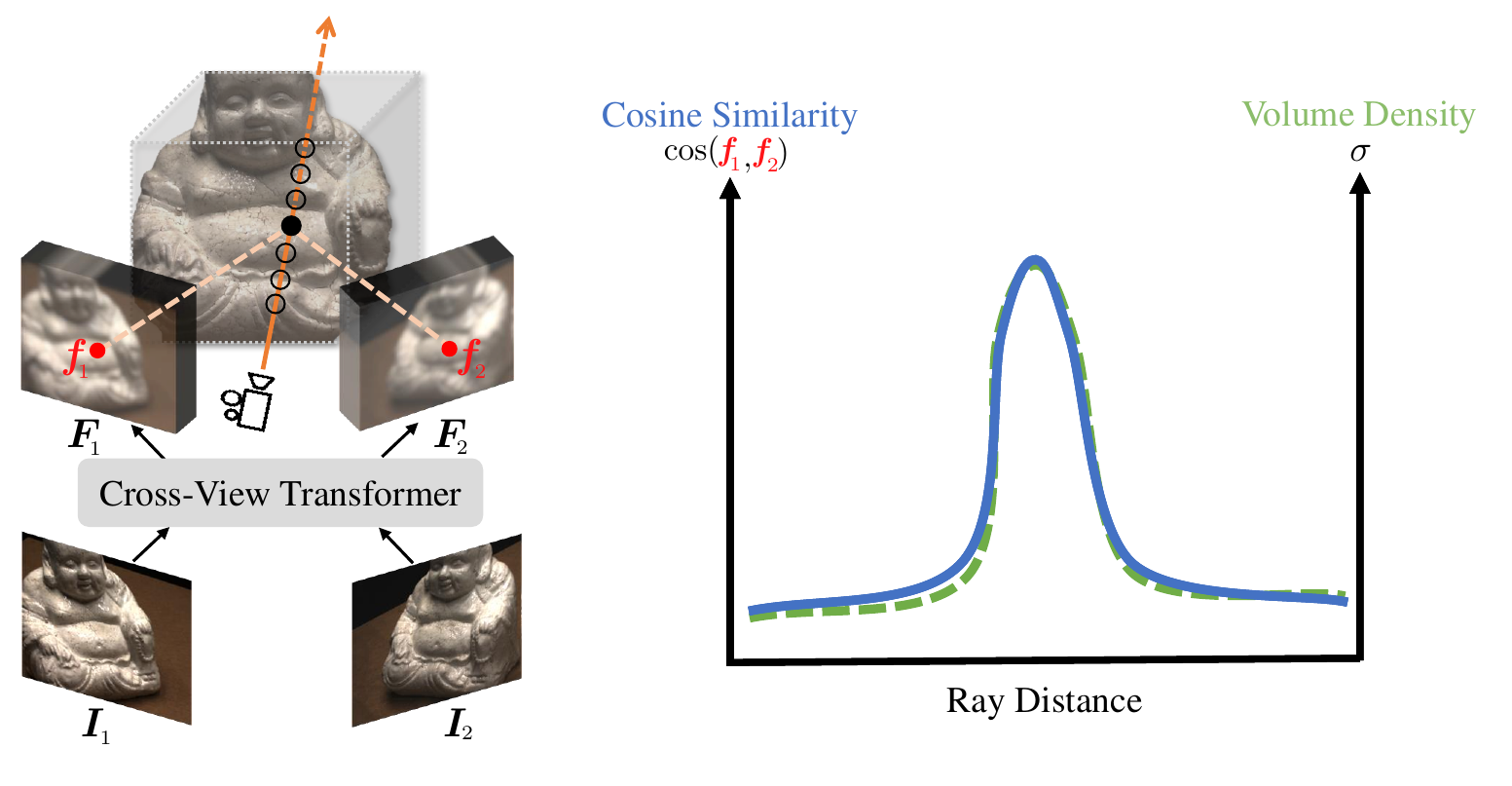}
    \end{center}
    \caption{\textbf{Correlation between cosine feature similarity and volume density}.
    We first extract image features via a Transformer by considering cross-view interactions.
    Then, we explicitly fetch the correspondence feature matching information by computing the cosine similarity between sampled features, which shows strong correlation with volume density and thus provides valuable geometric cues for density prediction. 
    }
    \label{fig:intro_idea}
\end{figure}

The limitations of cost-volume-based methods motivate us to turn back to the 2D image features for a simple and effective, generalizable NeRF alternative.
Our \emph{key} idea is to \emph{explicitly match 2D image features across different views and use the correspondence matching statistics as the geometry prior $\bm z$}.
This is meaningful because the feature matching (\ie, the cosine similarity) of 2D projections of a 3D point reflects the multi-view consistency, which encodes the scene geometry and correlates well with the volume density \red{for opaque and non-occluded surfaces} (see Fig.~\ref{fig:intro_idea}).
One \emph{key} observation we have is that the 2D image features need to be aligned across different views, which can be effectively done via Transformer cross-attention over pairs of views. 
Like our work, GPNR~\cite{suhail2022generalizable} uses 2D Transformer blocks to \emph{implicitly} obtain correspondence matching by aggregating features along the epipolar line and across different views, and then predicts pixel colors using the fused features.
In contrast, our method \emph{explicitly} computes feature correspondence matching as cosine similarity for the NeRF conditional inputs.
By providing NeRF with explicit geometry cues, our model manages to generalize to new scenes more effectively and efficiently (see TABLE~\ref{tab:ablations}\subref{tab:matching} and TABLE~\ref{tab:gpnr_sota}).

Specifically, we explicitly compute multi-view correspondence matching using features extracted via self- and cross-attention of a Transformer, which capture self- and cross-view interactions.
Our framework builds upon GMFlow~\cite{xu2022gmflow,xu2023unifying}, an off-the-shelf model originally designed for dense optical flow estimation, finding dense correspondence between two images.
In particular, we adapt its encoder to handle \emph{arbitrary views} for multi-view inputs by processing the features in a pair-wise manner.
Each pair of features is fed into the Transformer together to extract stronger and aligned features.
For each 3D point, we project it into different views based on the given camera parameters, and bilinearly sample the Transformer features.
We then compute the cosine similarity of sampled features in a pairwise manner.
To improve the expressiveness of cosine similarity, we compute it in a group-wise manner along the channel dimension,  similar to \cite{guo2019group,xu2020learning}. Finally, the averaged group-wise cosine similarity is fed into the NeRF decoder along with the original NeRF coordinate input to predict color and density.
We demonstrate the superiority of our proposed method by thorough ablations and extensive experiments under several different evaluation settings.

Our major contributions can be summarized as follows:
\begin{itemize}
\item We propose to explicitly match 2D image features across different views and use the correspondence feature matching statistics as the geometry prior in place of 3D cost volumes for generalizable NeRF.  
\item We implement explicit correspondence matching as the simple group-wise cosine similarity between image features, which are aligned via Transformer cross-attention to capture cross-view interactions.
\item Our method is view-agnostic, unlike popular cost volume-based approaches that are typically sensitive to reference view selection. 
\item Furthermore, we achieve state-of-the-art results on standard benchmarks including DTU~\cite{jensen2014large}, Real Forward-Facing (RFF)~\cite{mildenhall2020nerf} and Blender~\cite{mildenhall2020nerf}.

\end{itemize}

\minhl{
Our work establishes a core concept: \emph{feature matching} serves as a robust geometry prior for generalizable 3D scene reconstruction in a feed-forward manner.
While this work is initially explored within an implicit NeRF-based framework, it is fundamentally representation-agnostic.
Our subsequent work, MVSplat~\cite{chen2024mvsplat} (published in ECCV'24), demonstrates that the same geometry prior from \emph{feature matching} also significantly benefits feed-forward frameworks based on explicit representation, \ie, 3D Gaussian Splatting (3DGS)~\cite{kerbl20233d}.
This highlights the fundamental nature of our contribution in enhancing reconstruction fidelity across diverse 3D representations.
}

Remaining parts of the manuscript are organized as follows: Related NeRF-based novel view synthesis approaches and techniques will be discussed in Sec.~\ref{sec:related_work}.
We will then elaborate on the architecture and training of our model in Sec.~\ref{sec:method}.
Extensive ablation studies and comparisons with state-of-the-art approaches under various settings will be illustrated and analyzed in Sec.~\ref{sec:experiments}. Sec.~\ref{sec:conclusion} will conclude our work and discuss its limitations and potential future work.
Readers are referred to the project page for additional video results and to the released code for more technical details.

\section{Related Work}
\label{sec:related_work}
\subsection{Novel View Synthesis via NeRF}

With the powerful ability in modeling an implicit continuous 3D field and the volume rendering techniques~\cite{kajiya1984ray},
Neural Radiance Field (NeRF)~\cite{mildenhall2020nerf} paved a new way for solving novel view synthesis.
However, vanilla NeRF requires dense calibrated images and a lengthy time for per-scene optimization. Recent attempts have been made to make it more applicable, including reducing the number of images using additional regularization terms~\cite{Niemeyer2021Regnerf,truong2023sparf,deng2022depth,chen2022geoaug,yuan2022neural,yang2023freenerf,wang2023sparsenerf} and reducing the training time with better data structure~\cite{SunSC22,yu2021plenoctrees,muller2022instant,Chen2022ECCV,tang2022compressible}.

\begin{figure*}[t!]
    \begin{center}
    \includegraphics[width=0.99\linewidth]{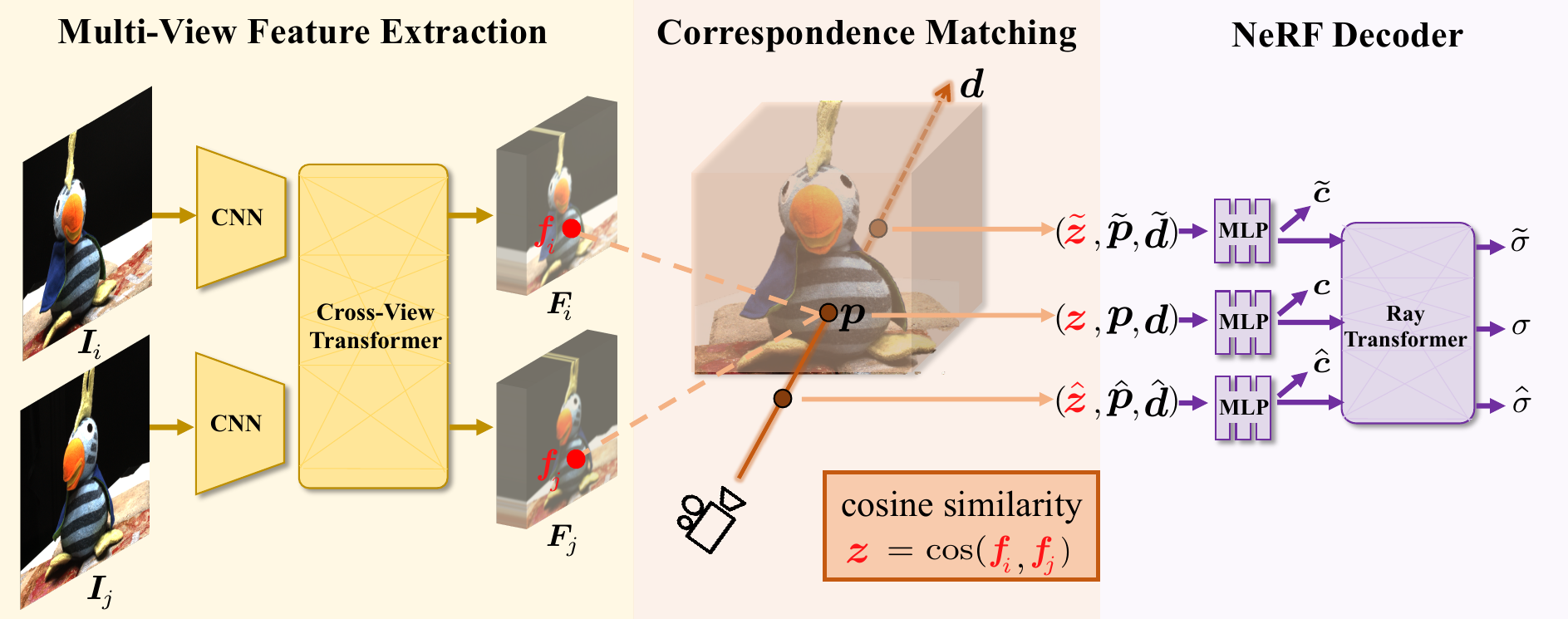}
    \end{center}
    \caption{\textbf{MatchNeRF overview}.
    Given $N$ input images, we extract the Transformer features and compute the cosine similarity in a pair-wise manner, and finally merge all pair-wise cosine similarities with element-wise average.
    \textbf{\rom{1})} For image pair ${\bm I}_i$ and ${\bm I}_j$, we first extract downsampled convolutional features with a weight-sharing CNN.
    The convolutional features are then fed into a Transformer to model cross-view interactions with cross-attention (Sec.~\ref{sec:feature}).
    \textbf{\rom{2})} To predict the color and volume density of a point on a ray for volume rendering, we project the 3D point into the 2D Transformer features ${\bm F}_i$ and ${\bm F}_j$ with the camera parameters and bilinearly sample the feature vectors ${\bm f}_i$ and ${\bm f}_j$ at the projected locations.
    We then compute the cosine similarity ${\bm z} = \cos({\bm f}_i, {\bm f}_j)$ between sampled features to encode the correspondence matching information (Sec.~\ref{sec:matching}).
    \textbf{\rom{3})} ${\bm z}$ is next used with the 3D position $\bm p$ and 2D view direction $\bm d$ for color $\bm c$ and density $\sigma$ prediction. An additional ray Transformer is used to model cross-point interactions along a ray (Sec.~\ref{sec:nerf_decoder}).}
    \label{fig:archi}
\end{figure*}

\subsection{Generalizable NeRF from a Learned Prior}

Pioneer works~\cite{yu2021pixelnerf,wang2021ibrnet,trevithick2021grf} resort to condition NeRF on 2D features extracted from each input view, which however results in poor visual quality for unseen scenes due to the lack of explicit geometry-aware encoding. Following approaches~\cite{chibane2021stereo,chen2021mvsnerf,johari2022geonerf,liu2022neural,xu2022point} demonstrate that introducing a geometric prior could improve the generalization. Specifically, the majority of them implement the prior by constructing a 3D cost volume with 3D CNNs for post-regularization, which is sensitive to the selection of the reference view. 
In contrast, we introduce a matching-based strategy to incorporate geometry prior without requiring the 3D cost volume nor the subsequent 3D CNN, and it is view-agnostic.
Several works~\cite{liu2023zero,tang2023make,chan2023generative,qian2023magic123,shi2023zero123++} have showcased promising results in single-view image-to-3D generation by integrating the diffusion model~\cite{saharia2022photorealistic} with NeRF. 
\red{Although addressing the same novel view synthesis task, our work shares minimal overlap with these diffusion-based approaches in terms of research focus and contribution. Diffusion-based methods primarily demonstrate strong generative capabilities, synthesizing plausible unseen regions from a \emph{single-view} input, but are often limited to \emph{object-level} scenes due to the difficulty of modeling complex data distributions. In contrast, our model focuses on leveraging geometric priors by explicitly matching correspondences across multiple views. This effective \emph{multi-view} fusion module enables it to handle more \emph{complex scenes} with multiple objects and complex backgrounds.}

\subsection{Transformer in NeRF} 

Recently, several methods have explored integrating the Transformer~\cite{vaswani2017attention} architecture into the NeRF models.
Notably, NerFormer~\cite{reizenstein2021common} employs a Transformer to aggregate features between different views on a ray and learns the radiance fields from the aggregated features. 
IBRNet~\cite{wang2021ibrnet} proposes a ray Transformer to aggregate the information on a ray in the NeRF decoder.
More recently, GNT~\cite{t2023is} directly regresses the color for view synthesis without NeRF's volume rendering.
GPNR~\cite{suhail2022generalizable} \red{extends the per-scene optimization method LFNR~\cite{suhail2022light} and} aggregates features with several stacked ``patch-based'' Transformers to improve generalization by avoiding potential harmful effects caused by CNN.
Unlike the above methods, \minhl{our approach primarily utilizes a Transformer-based encoder} to enhance the features through cross-view interactions, so that it can better serve the following correspondence matching process, aiming to provide a better geometry prior.

\section{Methodology}
\label{sec:method}
Our goal is to learn a \emph{generalizable} NeRF that synthesizes novel views for unseen scenes with \emph{few} (2 or 3) input views \minhl{in a single forward pass}.
This task is conceptually built on vanilla NeRF~\cite{mildenhall2020nerf}, except that here the model is trained for a series of scenes, rather than optimizing the model for each scene.
Once trained, the model can produce high-quality novel views for conditional inputs, \emph{without \minhl{additional training}}.
In particular, given $N$ views $\mathcal{I} = \{{\bm I}_i\}^N_{i=1}$ of a scene and corresponding camera parameters $\mathcal{M} = \{{\bm M}_i\}^N_{i=1}$, most generalizable NeRF approaches~\cite{chibane2021stereo,johari2022geonerf,wang2021ibrnet,chen2021mvsnerf} can be formulated as finding the functions of view fusion $f_{\theta}$ and view rendering  $g_{\phi}$,
\begin{equation} 
\label{eq:i2v_mapping}
f_{\theta}: (\mathcal{I}, \mathcal{M}) \to {\bm z}, \quad g_{\phi}: ({\bm p}, {\bm d}, {\bm z}) \to ({\bm c}, \sigma),
\end{equation}
where $\bm p$ and $\bm d$ are the 3D point position and the 2D ray direction of a target viewpoint, $\bm c$ and $\sigma$ are the predicted color and density, which are used to render the target novel view via volume rendering (Sec.~\ref{sec:nerf_decoder}) as in the vanilla NeRF model.
The key difference with vanilla NeRF is that additional information $\bm z$ is injected to provide the geometry prior, %
which is usually derived from images ($\mathcal{I}$ and $\mathcal{M}$).
The functions $f_{\theta}$ and $g_{\phi}$ are typically parameterized with deep neural networks, where $\theta$ and $\phi$ denote the learnable parameters of the networks.

Previous generalizable NeRF approaches~\cite{chibane2021stereo,johari2022geonerf,wang2021ibrnet,chen2021mvsnerf} \minhl{differ primarily} in how the \minhl{geometry prior} $\bm z$ is encoded with various networks $f_{\theta}$.
In this
\minhl{work, we introduce \textbf{MatchNeRF}, a new framework that} explicitly matches 2D image features across different views and uses the \minhl{resulting correspondence} statistics as a geometry prior $\bm z$. %
Fig.~\ref{fig:archi} gives an overview of our framework.
It consists of a Transformer encoder $f_{\theta}$ to extract cross-view enhanced features, an explicit cosine similarity computation operation to obtain the correspondence matching information for the geometry prior $\bm z$, and a NeRF decoder $g_{\phi}$  with a ray Transformer to predict the color and density for volume rendering.

\subsection{Multi-View Feature Extraction}
\label{sec:feature}

For $N$ input views $\{{\bm I}_i\}^N_{i=1}$, we first extract $8 \times$ downsampled convolutional features $\{{\bm F}^{\mathrm{c}}_i\}^N_{i=1}$ for each view \emph{independently}, using a weight-sharing CNN.
Unlike prior approaches~\cite{yu2021pixelnerf,wang2021ibrnet,chen2021mvsnerf} that are solely dependent on the per-view convolutional features, we further improve the feature quality by modeling the \emph{cross-view interactions} between different views.
This is achieved through a Transformer with \emph{cross-attention}, for which we build upon the GMFlow's~\cite{xu2022gmflow} Transformer architecture.
Compared with GMFlow, our proposed method can handle arbitrary views for novel view synthesis in a \emph{pair-wise} manner, instead of working only for fixed two views in the optical flow setting.

More specifically, we consider all the possible 2-view combinations of $N$ views and obtain a total of $N(N-1)/2$ view pairs, \red{similar to SRF~\cite{chibane2021stereo}. Although this quadratic computation might appear costly, it remains practical in our sparse-view setting, as we were able to run all experiments (2 to 6 views) on \emph{a single 16G-V100 GPU} (see Sec.~\ref{sec:exp_implement})}.
Each pair of convolutional features is fed into a weight-sharing Transformer \emph{jointly} to consider their cross-view interactions, which can be described as 
\begin{equation}
    \mathcal{T}: ({\bm F}^{\mathrm{c}}_i, {\bm F}^{\mathrm{c}}_j) \to ({\bm F}_i, {\bm F}_j), \quad \forall i, j \in \{1, 2, \cdots, N\}, i < j,
\end{equation}
where $\mathcal{T}$ denotes the Transformer, which is composed of six stacked Transformer blocks, where each Transformer block contains self-, cross-attention, and feed-forward networks following GMFlow~\cite{xu2022gmflow}.
We also add the fixed sine and cosine positional encodings to the convolutional features ${\bm F}^{\mathrm{c}}_i$ and ${\bm F}^{\mathrm{c}}_j$ before computing the attentions to inject the positional information~\cite{carion2020end}.
We compute the attentions in a shifted local window~\cite{liu2021swin} manner for better efficiency.

Thus far, the features extracted from the Transformer are at $1/8$ of the original image resolution, which is lower than the $1/4$ resolution used in MVSNeRF~\cite{chen2021mvsnerf}, may therefore limit performance.
To remedy this, we align our highest feature resolution with MVSNeRF by \minhl{using a lightweight} convolutional upsampler to do $2 \times$ upsampling on the $1/8$ features.
\minhl{As a result}, we obtain both $1/8$ and $1/4$ resolution features.
This simple upsampler leads to notable \minhl{performance gains}, as \minhl{demonstrated in our ablations in}  TABLE~\ref{tab:ablations}\subref{tab:feature_scales}.
We denote all the $1/4$ resolution feature pairs after upsampling as $(\hat{\bm F}_i, \hat{\bm F}_j), i, j \in \{1, 2, \cdots, N\}, i < j$.

\subsection{Correspondence Matching}
\label{sec:matching}

To further provide geometric cues for volume density prediction, we compute the cosine similarity between sampled Transformer features as the correspondence matching information.
Specifically, for a given 3D point position $\bm p$, we first project it onto the 2D Transformer features ${\bm F}_i$ and ${\bm F}_j$ ($i, j \in \{1, 2, \cdots, N\}, i < j$) of views $i, j$ using the camera parameters ${\bm M}_i$ and ${\bm M}_j$, and then bilinearly sample the corresponding feature vectors ${\bm f}_i$ and ${\bm f}_j$ at the specific 2D positions.
We compute the cosine similarity between sampled Transformer features ${\bm f}_i$ and ${\bm f}_j$ to measure the multi-view consistency so as to provide the NeRF decoder information about whether the 3D point $\bm p$ is on the surface or not.
However, the basic cosine similarity merges two high-dimensional feature vectors to a single scalar, which might lose too much information (see TABLE~\ref{tab:ablations}\subref{tab:matching}).
To further improve its expressiveness, we calculate the cosine similarity in a group-wise manner along the channel dimension~\cite{guo2019group}.
Specifically, the feature vectors ${\bm f}_i$ and ${\bm f}_j$ are first equally partitioned into $G$ groups, and we then compute the cosine similarity between features in each group:
\begin{equation}
    {s}^{(g)} = \frac{{\bm f}_i^{(g)} \cdot {\bm f}_j^{(g)} }{\left \| {\bm f}_i^{(g)} \right \|_2 \cdot  \left \| {\bm f}_j^{(g)} \right \|_2}, \quad g = 1, 2, \cdots, G,
\end{equation}
where $(g)$ denotes the $g$-th group. All the cosine similarities are collected as a vector $\bm s = ({s}^{(1)}, {s}^{(2)}, \cdots, {s}^{({G})}) \in \mathbb{R}^{{G}}$.

Recall that the Transformer features are extracted at both $1/8$ and $1/4$ resolutions. Similarly we compute group-wise cosine similarity between the $1/4$ resolution features $\hat{\bm F}_i$ and $\hat{\bm F}_j$, $i, j \in \{1, 2, \cdots, N\}, i < j$.
This yields another similarity $\hat{\bm s} = ({\hat{s}}^{(1)}, {\hat{s}}^{(2)}, \cdots, {\hat{s}}^{(\hat{G})}) \in \mathbb{R}^{\hat{G}}$ for $\hat{G}$ groups.
We then concatenate ${\bm s}$ and $\hat{\bm s}$ and obtain the cosine similarity ${\bm s}_{i,j} \in \mathbb{R}^{G+\hat{G}}$ for view pair $(i, j)$.

This process is repeated for all the view pairs $(i, j), i, j \in \{1, 2, \cdots, N\}, i < j$ for a total of $N$ views, and the final cosine similarity is obtained by taking an element-wise average over all $N(N-1)/2$ pairs:
\begin{equation}
\label{eq:avg_cosine}
    \bm z = \frac{\sum_{ij} {\bm s}_{ij}}{N(N-1)/2} \in \mathbb{R}^{G + \hat{G}},
\end{equation}
where $\bm z$ is the geometry prior of our method, capturing the explicit correspondence matching information.

\subsection{NeRF Decoder}\label{sec:nerf_decoder}

The cosine similarity $\bm z$ in Eq.~\eqref{eq:avg_cosine} constructed from the encoder $f_{\theta}$ is fed into the NeRF decoder $g_{\phi}$ for predicting NeRF color and density, as formulated in Eq.~\eqref{eq:i2v_mapping}.

\subsubsection{Rendering network}
We follow prior works~\cite{yu2021pixelnerf,chen2021mvsnerf} to construct a MLP-based rendering network.
Similarly, we also include the texture priors by concatenating the color information sampled on all input views with the given position.
But unlike the typical MLP decoder that processes all points on a ray independently, 
we further explore introducing \emph{cross-point interactions} by fusing the rendered information along a ray via a Transformer.
We adopt IBRNet's~\cite{wang2021ibrnet} ray Transformer in our implementation for convenience.

\subsubsection{Volume rendering}
With the emitted color $\bm{c}$ and volume density $\sigma$ predicted by the rendering network, novel views can be synthesized via volume rendering, which is implemented with differential ray marching as in NeRF~\cite{mildenhall2020nerf}.
Specifically, to estimate the color ${\bm{C}}$ of a pixel, radiance needs to be accumulated across all sampled shading points on the corresponding ray that passes through the pixel, 
\begin{equation}
    {\bm{C}} = \sum_{i=1}^{K} T_i(1-\textrm{exp}(-\sigma_i\delta_i))\bm{c}_i, \quad T_i = \textrm{exp}(-\sum_{j}^{i-1}\sigma_j\delta_j),
\end{equation}
where $\bm{c}_i, \sigma_i$ refer to the color and density of the $i$-th sampled 3D point on the ray.
$T_i$ is the volume transmittance, and $\delta_i$ denotes the distances between adjacent points.
$K$ is the total number of sampled 3D points on a ray.

\begin{table*}[t]
\caption{\textbf{MatchNeRF Ablations}. All models are trained and evaluated on the DTU dataset using the settings of 3 \emph{nearest} input views. Results are measured over only \emph{the foreground region} following the settings of MVSNeRF.}  \label{tab:ablations}
\vspace{-15pt}
\begin{center}
\subfloat[
{\footnotesize \textbf{Model components}. The baseline is a CNN-only feature extractor. ``self" and ``cross" denote the self- and cross-attention, and ``ray" denotes the ray Transformer. Cross-attention contributes the most.}
\label{tab:components}
]{
\begin{minipage}{0.45\linewidth}
{\begin{center}
\setlength{\tabcolsep}{5pt} %
\begin{tabular}{lccccccccccccccc}
    \toprule
    
    model & PSNR$\uparrow$ & SSIM$\uparrow$ & LPIPS$\downarrow$ \\
    \midrule
    
    CNN                      & 23.20 & 0.874 & 0.262  \\
    CNN + self               & 23.51 & 0.878 & 0.254   \\
    CNN + cross              & 26.13 & 0.922 & 0.184   \\
    CNN + self + cross       & 26.76 & 0.929 & 0.168   \\
    CNN + self + cross + ray & \textbf{26.91} & \textbf{0.934} & \textbf{0.159}   \\
    \bottomrule
    \end{tabular}
\end{center}}
\end{minipage}
}
\hspace{2em}
\subfloat[
{\footnotesize \textbf{Feature relation measures}. ``learned similarity" denotes using a network to \emph{implicitly} learn the similarity as introduced in SRF~\cite{chibane2021stereo}, where the network takes two features as input and outputs a scalar value.}
\label{tab:matching}
]{
\begin{minipage}{0.45\linewidth}{\begin{center}
\begin{tabular}{lccccccccccccccc}
    \toprule
    
    relation & PSNR$\uparrow$ & SSIM$\uparrow$ & LPIPS$\downarrow$ \\
    \midrule
    
    concatenation     &  24.21  &  0.893  &  0.219 \\
    learned similarity    &  24.87  &  0.906  &  0.206 \\
    variance          &  26.36  &  0.929  &  0.167 \\
    cosine            &  26.24  &  0.927  &  0.175 \\
    group-wise cosine &  \textbf{26.91}  &  \textbf{0.934}  &  \textbf{0.159} \\
    \bottomrule
    \end{tabular}
\end{center}}\end{minipage}
}
\\
\subfloat[
{\footnotesize \textbf{Number of Transformer blocks}. A single block can bring clear gains, and more blocks improve the performance further. }
\label{tab:blocks}
]{
\begin{minipage}{0.45\linewidth}{\begin{center}
\begin{tabular}{cccccccccccccccc}
    \toprule
    
    \#blocks & PSNR$\uparrow$ & SSIM$\uparrow$ & LPIPS$\downarrow$ \\
    \midrule
    
    0 &         23.20  &  0.874  &  0.262   \\
    1 &         25.35  &  0.910  &  0.204   \\
    3 &         26.38  &  0.925  &  0.180  \\
    6 &         \textbf{26.76}  &  \textbf{0.929}  &  \textbf{0.168} \\
    \bottomrule
    \end{tabular}
\end{center}}\end{minipage}
}
\hspace{2em}
\subfloat[
{\footnotesize \textbf{Feature resolution}. Combining both $1/8$ and $1/4$ resolution features leads to the best results.}
\label{tab:feature_scales}
]{
\begin{minipage}{0.45\linewidth}{\begin{center}
\begin{tabular}{ccccccccccccccc}
    \toprule
    
    feature resolution & PSNR$\uparrow$ & SSIM$\uparrow$ & LPIPS$\downarrow$ \\
    \midrule
    
    $1/8$          &   26.09  &  0.920  &  0.187  \\
    $1/4$          &   26.71  &  0.931  &  0.164  \\
    $1/8$ \& $1/4$ &   \textbf{26.91}  &  \textbf{0.934}  &  \textbf{0.159}  \\
    
    \bottomrule

    \\
    \end{tabular}
\end{center}}\end{minipage}
}
\\
\end{center}
\end{table*}

\subsection{Training Loss}\label{sec:loss}
Our full model is trained end-to-end with \emph{only} the photometric loss function, without requiring any ground-truth geometry data.In particular, we optimize the model parameters with the squared error between the rendered pixel colors and the corresponding ground-truth ones,
\begin{equation}
    \mathcal{L} = \sum_{p \in \mathcal{P}}  \left \|  {\bm{C}}_p - \tilde{\bm{C}}_p \right \|_2^2,
\end{equation}
where $\mathcal{P}$ denotes the set of pixels within one training batch, and $\bm{C}_p, \tilde{\bm{C}}_p$ refer to the rendered color and the ground-truth color of pixel $p$, respectively.

\minhl{
\subsection{Relationship to MVSplat}
The core concept of using feature matching as a geometry prior, introduced in MatchNeRF, is not limited to implicit NeRF-based frameworks.
Our subsequent work, MVSplat~\cite{chen2024mvsplat}, investigates the same concept in the context of the explicit 3DGS~\cite{kerbl20233d} representation.
MVSplat utilizes plane sweeping to perform feature matching for accurate localization of 3D Gaussian centers in a single forward pass, leading to improved visual and geometric quality as well as stronger generalization capability.
Although MatchNeRF and MVSplat adopt different 3D representations (implicit NeRF \vs explicit 3DGS), they are \emph{unified} by the same core contribution: cross-view feature matching provides a robust geometry cue for feed-forward 3D scene reconstruction.
} %

\section{Experiments}\label{sec:experiments}

\subsection{Experiment Settings}

\subsubsection{Datasets and evaluation settings}
We mainly follow the settings of  MVSNeRF~\cite{chen2021mvsnerf} to conduct novel view synthesis using 3 input views.
We also report results on a more challenging scenario with only 2 input views, which is the minimal number of views required for our and cost volume-based methods.
In particular, we train MatchNeRF with 88 scenes from DTU~\cite{jensen2014large}, where each scene contains 49 views with a resolution of $512 \times 640$.
The trained model is first tested on \minhl{the unseen} 16 scenes from DTU, then tested directly (without any fine-tuning) on 8 scenes from Real Forward-Facing (RFF)~\cite{mildenhall2020nerf} and 8 scenes from Blender~\cite{mildenhall2020nerf}, both of which contain significantly different contents and view distributions from DTU.
The resolutions of RFF and Blender are $640\times 960$ and $800 \times 800$, respectively.
Each test scene is measured with 4 novel views.
The performance is measured with PSNR, SSIM~\cite{wang2004image} and LPIPS~\cite{zhang2018unreasonable} metrics.

\subsubsection{Implementation details \label{sec:exp_implement}}
We initialize our feature extractor with GMFlow's pretrained weights, which saves us time in training our full model (\emph{further discussed in Appendix~\ref{sec:app_backbone}}).
We borrow the upsampler network from \cite{niemeyer2021giraffe}. 
The number of groups for group-wise cosine similarity is $G+\hat{G}=2+8$ (for $1/8$ and $1/4$ resolution features, respectively), chosen empirically to balance the dimension of similarity with that of the concatenated colors (3 $\times$ 3 for 3 views).
Our NeRF decoder includes an MLP that is in general the same as MVSNeRF and a lightweight ray Transformer.
128 points are uniformly sampled from each rendering ray, and only one single radiance field is reconstructed \emph{without} using the coarse-to-fine technique.
512 rays are randomly sampled in each training batch.

The model is trained with AdamW~\cite{loshchilov2017decoupled} optimizer, and the learning rates are initialized as 5e-5 for the encoder and 5e-4 for the decoder, decayed using the one cycle policy~\cite{smith2019super}.
We also clip the global norm of gradients of the cross-view Transformer to $\le 1$ to avoid exploding gradients.
Our MatchNeRF is implemented with PyTorch~\cite{paszke2019pytorch} and its default model is trained for 7 epochs, which takes around 28 hours on a single 16G-V100 GPU using the 3 input views settings.
Code and pretrained weights are available at \rurl{github.com/donydchen/matchnerf}.

\begin{figure*}[t]
    \begin{center}
    \includegraphics[width=\linewidth]{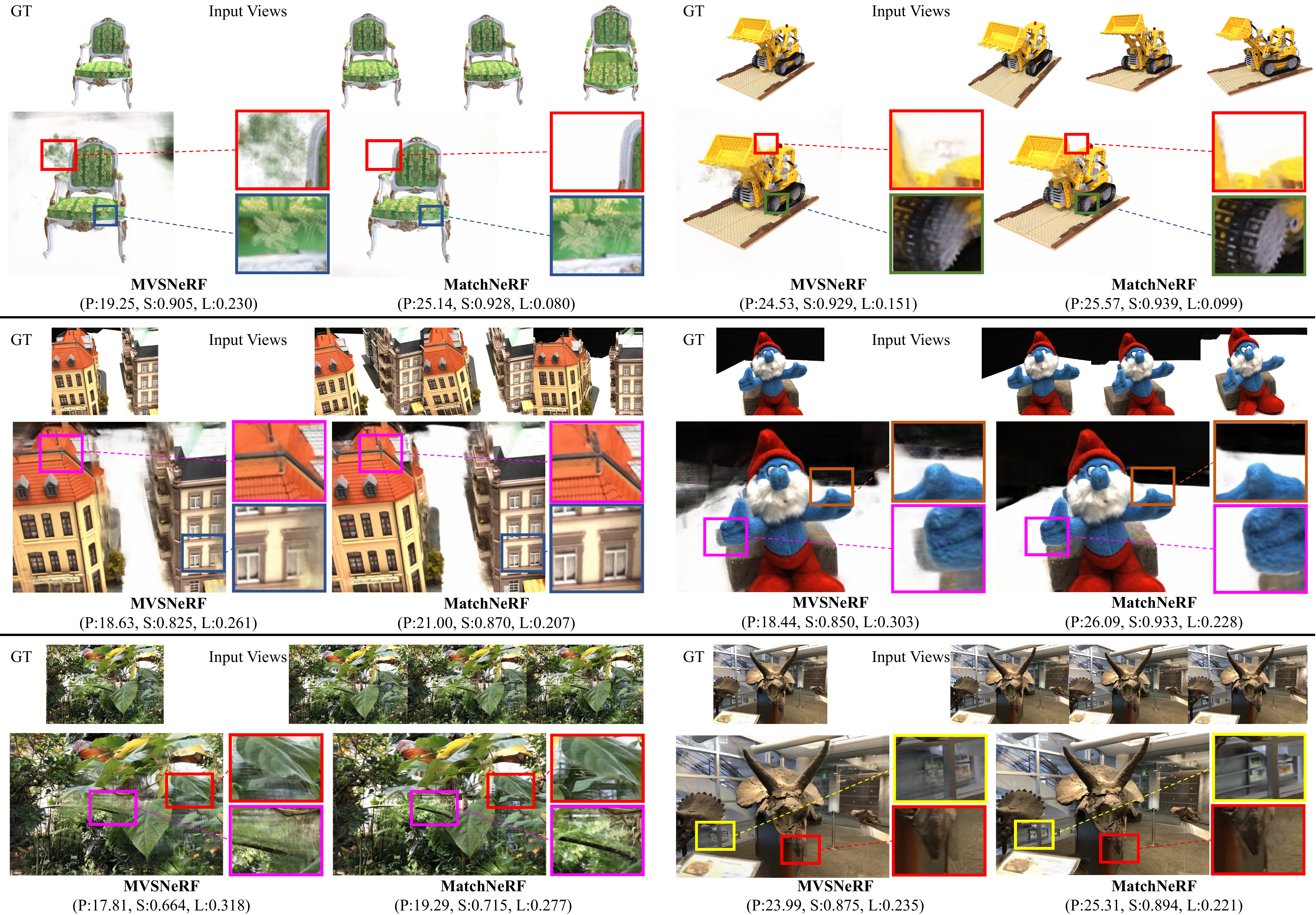}
    \end{center}
    \caption{\textbf{Qualitative results on Blender (1st row), DTU (2nd row) and RFF (3rd row)}.
    We showcase the visual results of MVSNeRF and our MatchNeRF method.
    Input views contain 3 viewpoints \emph{nearest} to the target one, and the first input view is the reference view for MVSNeRF.
    Our MatchNeRF reconstructs better details (`leaves' scene of RFF) and contains less background artifacts (`doll' scene of DTU).
    The construction of cost volume in MVSNeRF requires all other views to be warped to the reference view, which results in poor quality when some views are clearly different from the reference view (`chair' scene of Blender, \emph{elaborated in Appendix~\ref{sec:app_cv_limit}}).
    Quantitative results measured over \emph{the whole image} are placed below each image, which in order are scores of PSNR, SSIM and LPIPS.
    }
    \label{fig:comp}
\end{figure*}

\subsection{Ablation Studies}

\subsubsection{Model components} 
We evaluate the importance of different network components of our full model, reported in TABLE~\ref{tab:ablations}\subref{tab:components}.
We start from a baseline with only CNN-based feature extractor (`CNN'), which is commonly used in prior works~\cite{shin2018pixels,chibane2021stereo,wang2021ibrnet,chen2021mvsnerf}.
\minhl{Adding a Transformer with only self-attention (`CNN + self') results in only marginal performance improvement.}
However, the model is significantly improved when the backbone features are refined with a cross-attention component (`CNN+cross'), showing that it is critical to model \emph{cross-view interactions} in feature extraction.
We then use both self- and cross-attention to further enhance our backbone (`CNN+self+cross'). %
Finally, we add a ray Transformer to our decoder to enhance the \emph{cross-point interactions} along the ray space (`CNN+self+cross+ray'), which brings additional improvements. %

\subsubsection{Feature relation measures}
Another key component of our method is the explicitly modeled correspondence matching information, which is implemented with group-wise cosine similarity between sampled Transformer features.
We compare \minhl{our method against several potential counterparts in TABLE~\ref{tab:ablations}\subref{tab:matching} to highlight its effectiveness.}
A straightforward option would be directly concatenating the features.
However, \minhl{even though it leverages Transformer features as in the default model, this variant performs notably worse.
This suggests that stronger features alone are insufficient} in providing direct guidance to the density prediction.
SRF~\cite{chibane2021stereo} proposes to \emph{implicitly} learn the feature similarity with a network, instead of \emph{explicitly} computing the similarity with the parameter-free cosine function as in our method.
We implement SRF's method by predicting a scalar output from two input features as the implicitly learned similarity.
We can observe SRF's approach is considerably worse, suggesting that our explicit cosine feature similarity is more effective.
Next, we compare with the variance metric typically used in cost-volume-based approaches like MVSNeRF~\cite{chen2021mvsnerf}, our group-wise cosine similarity again performs better.
Compared to the basic cosine similarity that reduces two high dimensional features to a single scalar, which may lose important information, our group-wise computation is more advantageous and thus leads to the best results.

\begin{table*}[t]
\caption{\textbf{Comparison with SOTA Methods}. MatchNeRF performs the best for both 3- and 2-view inputs. Viewpoints \emph{nearest} to the target one are selected as input. By default we measure over only \emph{the foreground or central regions} following MVSNeRF's settings, while \textsuperscript{$\top$} indicates a more accurate metric by measuring over \emph{the whole image}. GeoNeRF is retrained without depth supervision for fair comparison.We measure MVSNeRF's 3-view whole image results with its pretrained weight, and retrain with its released code to report 2-view results. }\label{tab:sota}
\vspace{-5pt}
    \begin{center}
    \begin{tabular}{llccccccccccccccccccccccc}
    \toprule
    \multirow{2}{*}[-2pt]{Input} & \multirow{2}{*}[-2pt]{Method} & \multicolumn{3}{c}{DTU~\cite{jensen2014large}} & \multicolumn{3}{c}{Real Forward-Facing~\cite{mildenhall2020nerf}} & \multicolumn{3}{c}{Blender~\cite{mildenhall2020nerf}} \\
    \addlinespace[-10pt] \\
    \cmidrule(lr){3-5} \cmidrule(lr){6-8} \cmidrule(lr){9-11}
    \addlinespace[-10pt] \\
    & & PSNR$\uparrow$ & SSIM$\uparrow$ & LPIPS$\downarrow$ & PSNR$\uparrow$ & SSIM$\uparrow$ & LPIPS$\downarrow$ & PSNR$\uparrow$ & SSIM$\uparrow$ & LPIPS$\downarrow$  \\
    
    \midrule

    \multirow{6}{*}[-4pt]{3-view} & PixelNeRF~\cite{yu2021pixelnerf}                                     & 19.31          & 0.789          & 0.382             & 11.24          & 0.486          & 0.671             & 7.39           & 0.658          & 0.411              \\
    
    & SRF~\cite{chibane2021stereo}                                        & 22.12          & 0.845         & 0.292             & 17.36          & 0.628          & 0.442             & 18.77          & 0.813          & 0.305     \\
    & IBRNet~\cite{wang2021ibrnet}                                        & 26.04          & 0.917          & 0.190             & 21.79          & 0.786          & 0.279             & 22.44          & 0.874          & 0.195              \\
    & GeoNeRF\textsuperscript{*}~\cite{johari2022geonerf}                                        & 26.76          & 0.893        & \textbf{0.150}             & 22.06          & 0.740          & 0.249             & 22.73          & 0.864          & 0.182     \\
    & MVSNeRF~\cite{chen2021mvsnerf}                                       & 26.63  & 0.931  & 0.168     & 21.93  & 0.795  & 0.252     & \textbf{23.62} & \textbf{0.897} & 0.176      \\
    & MatchNeRF                              & \textbf{26.91} & \textbf{0.934} & 0.159    & \textbf{22.43} & \textbf{0.805} & \textbf{0.244}    & 23.20  & \textbf{0.897}  & \textbf{0.164}     \\

    \addlinespace[-10pt] \\
    \cmidrule(lr){2-11}
    \addlinespace[-10pt] \\
    
    & MVSNeRF\textsuperscript{$\top$}\cite{chen2021mvsnerf}                                       & 20.67 & 0.865 & 0.296 & 20.78 & 0.778 & 0.283 & 24.63 & 0.929 & 0.155     \\
    & MatchNeRF\textsuperscript{$\top$} & \textbf{24.54} & \textbf{0.897} & \textbf{0.257} & \textbf{21.77} & \textbf{0.795} & \textbf{0.276} & \textbf{24.65} & \textbf{0.930} &  \textbf{0.120}                                \\

    \midrule
    \multirow{4}{*}[-1pt]{2-view} & MVSNeRF~\cite{chen2021mvsnerf} & 24.03 & 0.914 & 0.192 & 20.22 & 0.763 & 0.287 & 20.56 & 0.856 & 0.243 \\

    & MatchNeRF  & \textbf{25.03} & \textbf{0.919} & \textbf{0.181} & \textbf{20.59} & \textbf{0.775} & \textbf{0.276} & \textbf{20.57} & \textbf{0.864} & \textbf{0.200} \\

    \addlinespace[-10pt] \\
    \cmidrule(lr){2-11}
    \addlinespace[-10pt] \\

    & MVSNeRF\textsuperscript{$\top$}\cite{chen2021mvsnerf} & 19.08 & 0.831 & 0.340 & 18.13 & 0.730 & 0.345 & 21.68 & 0.875 & 0.216 \\
    & MatchNeRF\textsuperscript{$\top$}   & \textbf{23.66} & \textbf{0.886} & \textbf{0.285} & \textbf{19.28} & \textbf{0.760} & \textbf{0.310} & \textbf{22.11} & \textbf{0.906} & \textbf{0.151} \\
    
    \bottomrule
    \end{tabular}
    \end{center}
\end{table*}

\subsubsection{Number of Transformer blocks}
We also verify the gain of stacking different numbers of Transformer blocks for the encoder.
Our default model uses a stack of 6 Transformer blocks, where each block contains a self-attention and a cross-attention layer.
TABLE~\ref{tab:ablations}\subref{tab:blocks} shows the results of stacking 0, 1, 3, 6 Transformer blocks, respectively.
Our model observes notable improvement when augmented with 1 Transformer block.
It sees better performance when stacked with more blocks, and reaches its best with 6 blocks.

\subsubsection{Feature resolution}
We extract features at both $1/8$ and $1/4$ resolutions, and we compare their effects in TABLE~\ref{tab:ablations}\subref{tab:feature_scales}.
When comparing between $1/8$ and $1/4$ feature resolutions, 
we observe that the $1/4$ resolution model performs better, especially in terms of SSIM and LPIPS metrics.
Combining both $1/8$ and $1/4$ resolution features leads to the best performance and thus is used in our final model.

\subsection{Main Generalizable Comparisons}

\subsubsection{SOTA comparisons}
We mainly compare MatchNeRF with state-of-the-art generalizable NeRF models, namely PixelNeRF~\cite{yu2021pixelnerf}, SRF~\cite{chibane2021stereo}, IBRNet~\cite{wang2021ibrnet}, GeoNeRF~\cite{johari2022geonerf} (retrained without depth supervision) and MVSNeRF~\cite{chen2021mvsnerf}, under few-view settings (3- and 2-view).
Generally, many MVSNeRF follow-up works share similar limitations, as they all rely on cost volume.
We thus choose MVSNeRF, one of the most representative cost volume-based methods, as our main comparison to showcase the potential of our matching-based design.

\begin{figure}[t]
    \begin{center}
    \includegraphics[width=\linewidth]{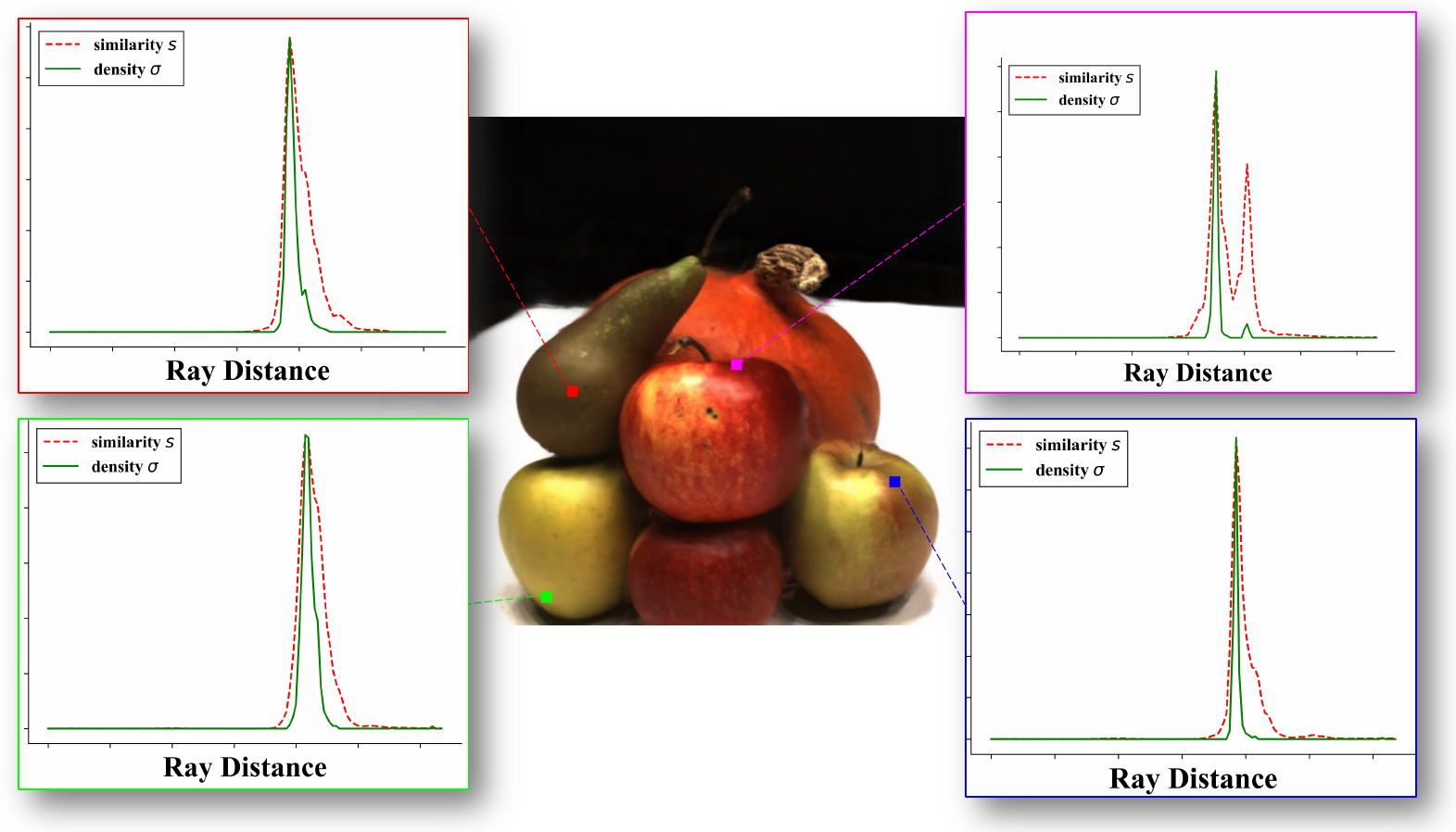}
    \end{center}
    \caption{
    \textbf{Relationship between the learned cosine similarity and volume density}. Four pixels are \emph{randomly selected} from the foreground of a DTU test scene (`scan63'). For each pixel, we showcase the learned cosine similarity (scalar value, predicted by the ablation model `cosine' in TABLE~\ref{tab:ablations}%
    ) and volume density of all sampled points along the corresponding ray. 
    The strong correlation demonstrates that our proposed cosine similarity is able to provide valuable geometric cues for volume density prediction.
    }
    \label{fig:cos2den}
\end{figure}

Quantitative results in TABLE~\ref{tab:sota} show that our MatchNeRF performs the best in all datasets under the settings of both 3- and 2-view.
\emph{By default we measure over only the foreground (DTU) or central (RFF and Blender) regions following the settings of MVSNeRF.}
Such settings are mainly introduced to ignore the background, as prior works tend to focus on the quality of foreground objects or scenes within small volumes, mostly due to the limitations of cost volume.
Considering that the reconstruction of entire scenes is more practical and is drawn increasing attention recently~\cite{reiser2023merf,tancik2022block}, it is undoubted that the background should also be included in measuring image quality, and thus \emph{it is more accurate to measure over the whole image}.
When reporting over the whole image (marked with \textsuperscript{$\top$} in TABLE~\ref{tab:sota}), the advantage of MatchNeRF is glaringly obvious, showing that our method can render high-quality contents for both the foreground and background.
Such a superiority can be more vividly observed from the visual results (Fig.~\ref{fig:comp}).
Notice that views rendered by MVSNeRF tend to contain artifacts around the background, since its cost volume is built towards one specific reference view, where the camera frustum might not have enough coverage of the target view.
Moreover, under the more challenging 2-view setting, MatchNeRF outperforms MVSNeRF by a larger margin, showing that our explicit correspondence matching provides more effective geometric cues for learning generalizable NeRF.

The main reason for the superiority of our model is that our encoder provides a compact and effective hint about the geometry structure for the decoder, making the learning process easier. 
Although MatchNeRF operates in 2D feature space similar to PixelNeRF, SRF and IBRNet, it significantly outperforms these methods, suggesting that our proposed cross-view-aware features do help the 3D estimation task.
Our method also outperforms GeoNeRF and MVSNeRF, which rely on 3D CNNs, indicating the effectiveness of our explicitly modeled correspondence matching information.
Fig.~\ref{fig:cos2den} further illustrates the strong correlation between the learned cosine similarity and volume density.

\begin{table}[t]
\caption{\textbf{Depth Reconstruction Comparisons on DTU}. `Abs error' and `Acc (0.05)' are short for absolute error and accuracy with a 0.05 threshold, respectively. Results are measured over \emph{the foreground} as DTU has \emph{no} ground-truth depth data for the background.\label{tab:depth}} 
\begin{center}
\begin{tabular}{lcccc}
\toprule
Metric                       & PixelNeRF & IBRNet & MVSNeRF & MatchNeRF       \\
\midrule
Abs error$\downarrow$ & 0.239     & 1.62   & 0.035   & \textbf{0.032}  \\
\midrule
Acc (0.05)$\uparrow$ & 0.187     & 0.001  & 0.866   & \textbf{0.886} \\
\bottomrule
\end{tabular}
\end{center}
\end{table}

\begin{figure}[t]
    \begin{center}
    \includegraphics[width=\linewidth]{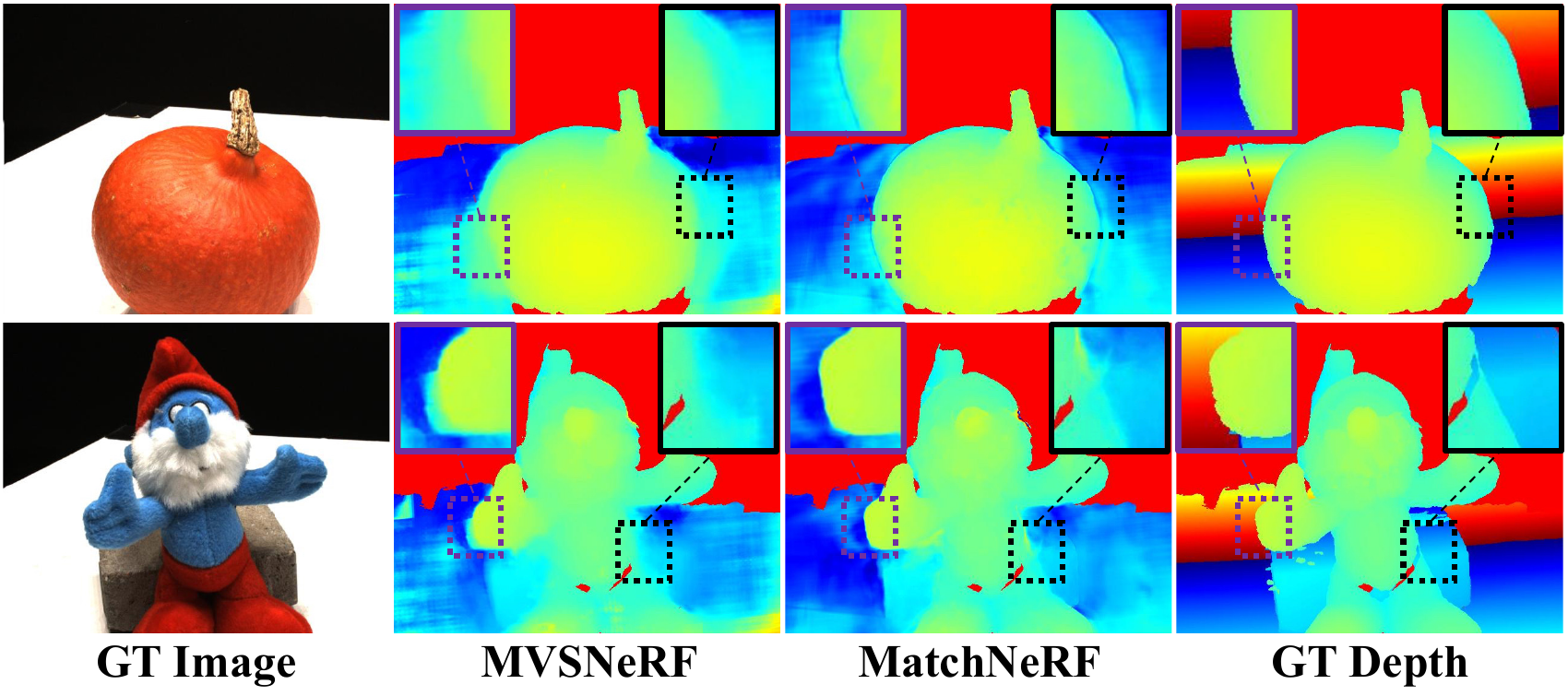}
    \end{center}
    \vspace{-5pt}
    \caption{\textbf{Visual results of rendered depth maps on DTU}. Background regions are masked out for depth maps rendered by both methods since ground-truth values are \emph{not} available for those regions. MatchNeRF reconstructs better depth with sharper borders.
    }
    \label{fig:depth_comp}
\end{figure}

\begin{figure*}[t]
    \begin{center}
    \includegraphics[width=\linewidth]{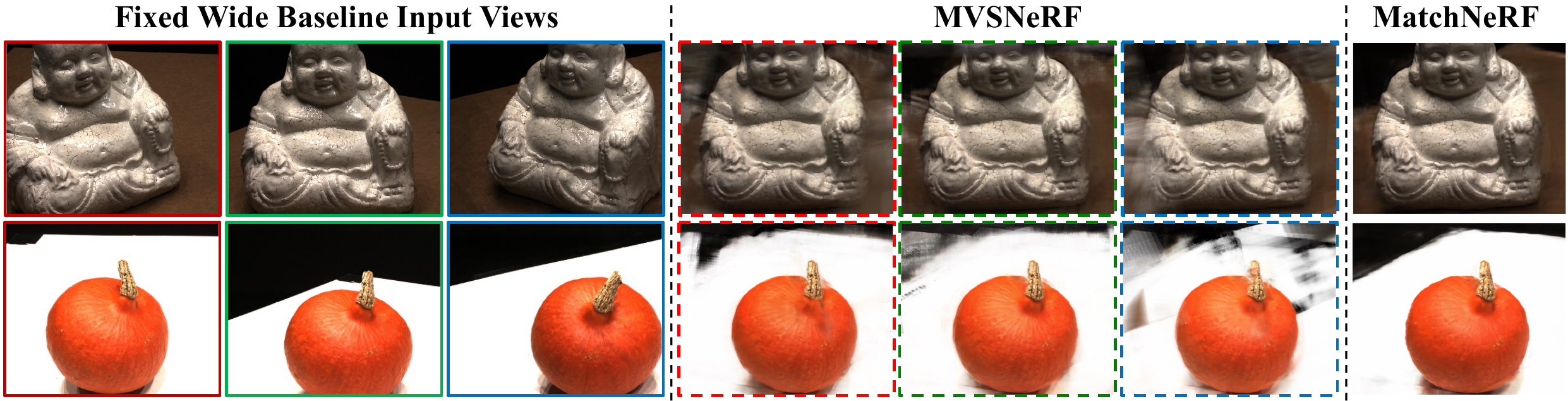}
    \end{center}
    \vspace{-10pt}
    \caption{\textbf{Visual results of using different reference views on DTU}. Three \emph{fixed wide baseline} views are used as input views. Each row presents the results of one target viewpoint from the corresponding test scene, and each image of MVSNeRF is rendered using the image with the same color border as the reference view. The visual quality of MVSNeRF becomes worse when the reference viewpoint is more different from the target one, especially around the background area. In contrast, our MatchNeRF is agnostic to the reference view.  
    }
    \vspace{-10pt}
    \label{fig:refviews}
\end{figure*}

\begin{table}[t]
\caption{\textbf{Effect of Selecting Different Reference Views on DTU}. Three \emph{fixed wide baseline} views are used as input views. `Ref.' states ID of the selected reference view. Results measured over \emph{the whole image} show that MVSNeRF is sensitive to the selection of reference view, while our MatchNeRF is view-agnostic.}
\label{tab:refviews}
\vspace{-5pt}
\begin{center}
\resizebox{0.95\linewidth}{!}{%
\begin{tabular}{lcccc} 
\toprule
Method                & Ref. ~ & PSNR$\uparrow$ & SSIM$\uparrow$ & LPIPS$\downarrow$  \\ 
\midrule
\multirow{3}{*}{MVSNeRF\textsuperscript{$\top$}\cite{chen2021mvsnerf}} & 1\textsuperscript{st}         &   16.98        &  0.766         &      0.391        \\
                         & 2\textsuperscript{nd}         &   17.87        &  0.787         &      0.375         \\
                         & 3\textsuperscript{rd}         &   14.92        &  0.715         &      0.434         \\ 
\midrule
MatchNeRF\textsuperscript{$\top$}                & -      &   \textbf{18.65}        &  \textbf{0.818}        &      \textbf{0.339}       \\
\bottomrule
\end{tabular}
}
\end{center}
\end{table}

\subsubsection{Depth reconstruction}\label{sec:app_depth}
We compare with state-of-the-art generalizable NeRF models regarding depth reconstruction (see TABLE~\ref{tab:depth}), our MatchNeRF again achieves the best results.
The main reason is that the depth map is rendered from the volume density (as it is done in NeRF~\cite{mildenhall2020nerf}), and our matching-based encoder can provide reliable geometry cues for density prediction (see Fig.~\ref{fig:cos2den}), thus leading to better reconstructed depth.
Visual results presented in Fig.~\ref{fig:depth_comp} further reassure our findings, notice that MatchNeRF reconstructs sharper and cleaner borders compared to MVSNeRF.

\begin{table}[t]
\caption{\textbf{MatchNeRF with Various Input Views on DTU}. MatchNeRF can perform better when given more input views, but it also takes a longer runtime. The results are reported on DTU using nearest input views, measured over \emph{the foreground region}. Runtime is measured over a test batch (\emph{4096-rays}). \label{tab:moreviews}} 
\begin{center}
\resizebox{0.95\linewidth}{!}
{%
\begin{tabular}{cccccc} 
\toprule
Input & PSNR$\uparrow$ & SSIM$\uparrow$ & LPIPS$\downarrow$  & Runtime$\downarrow$  \\ 

\midrule
2-view       & 25.03 & 0.919 & 0.181  & \textbf{0.06 s} \\
3-view       & 26.91          & 0.934          & 0.159       &  0.08 s   \\
4-view       & 27.04          & 0.936          & 0.159       &  0.11 s    \\
5-view       & 27.11          & 0.936          & 0.161       &  0.15 s    \\
6-view       & \textbf{27.38} & \textbf{0.938} & \textbf{0.157}  & 0.20 s \\ 
\bottomrule
\end{tabular}
}
\end{center}
\end{table}

\begin{table}[t]
\caption{\textbf{Comparisons on RFF using `GPNR Setting 1'}. Despite using only 3 input views, MatchNeRF performs the best compared to other methods that require 5 or even 10 input views. \label{tab:gpnr_sota}}
\vspace{-15pt}
\begin{center}
\resizebox{\linewidth}{!}{%
\begin{tabular}{lcccc} 
\toprule
Method                & Input & PSNR$\uparrow$ & SSIM$\uparrow$ & LPIPS$\downarrow$  \\ 
\midrule

IBRNet~\cite{wang2021ibrnet} & 10-view & 24.33 & 0.801 & \textbf{0.213} \\
\midrule
\multirow{2}{*}{GPNR~\cite{suhail2022generalizable}} & 5-view         &   24.33        &  0.850         &      0.216        \\
                         & 3-view         &   22.36        &  0.800         &      0.286         \\
\midrule
MatchNeRF                & 3-view      &   \textbf{24.82}        &  \textbf{0.860}        &      0.214       \\
\bottomrule
\end{tabular} 
}
\end{center}
\end{table}

\subsubsection{Effect of reference view selection}
The above experiments follow MVSNeRF's setting of choosing input views \emph{nearest} to the target one, which again helps ``hide'' the limitations caused by the cost volume.
Here we re-evaluate the models by using three \emph{fixed wide baseline} input views to further demonstrate the superiority of our design. 
As shown in Fig.~\ref{fig:refviews}, the performance of MVSNeRF is sensitive to the selection of reference view, where the performance drops significantly when the selected reference view does not have sufficient overlap with the target one.
In contrast, our MatchNeRF is by design view-agnostic, and achieves good performance even for background regions.
Quantitative results (see TABLE~\ref{tab:refviews}) measured over the whole image on DTU test scenes further reassure our findings.

\subsubsection{Effect of various number of input views}\label{sec:app_moreview}
Our MatchNeRF directly uses all input views by calculating matching similarity on pair-wise cross-view-aware features, thus it is by design able to capture more information when given more input views.
As reported in TABLE~\ref{tab:moreviews}, our MatchNeRF indeed achieves better performance with more input views.
Besides, we also report the average inference time for each test batch (contains 4096 rays in our experiment).
As expected, the runtime slightly increases as the number of views grows.

\subsubsection{Comparisons on GPNR Setting 1}
For fair comparison with prior work, we also evaluate MatchNeRF under another commonly used setting.
Following `GPNR Setting 1', we train on LLFF~\cite{mildenhall2019local} and IBRNet~\cite{wang2021ibrnet} collected scenes, and conduct generalization test on RFF~\cite{mildenhall2020nerf} dataset.
Readers are referred to~\cite{suhail2022generalizable} for more setting details.

As reported in TABLE~\ref{tab:gpnr_sota}, our MatchNeRF with only 3 input views is able to outperform IBRNet~\cite{wang2021ibrnet} with 10 input views in terms of PSNR and SSIM.
MatchNeRF also showcases its superiority by achieving better performance than 3-view and even 5-view GPNR~\cite{suhail2022generalizable}.
Note that  GPNR~\cite{suhail2022generalizable} is trained on 32 TPUs whereas MatchNeRF is trained on only one single 16G-V100 GPU.
This reassures us that \emph{explicit} correspondence matching can provide reliable geometry cues, enabling generalizable NeRF models to generalize to unseen scenes in a more effective and efficient manner.

\begin{figure}[t]
    \begin{center}
    \includegraphics[width=\linewidth]{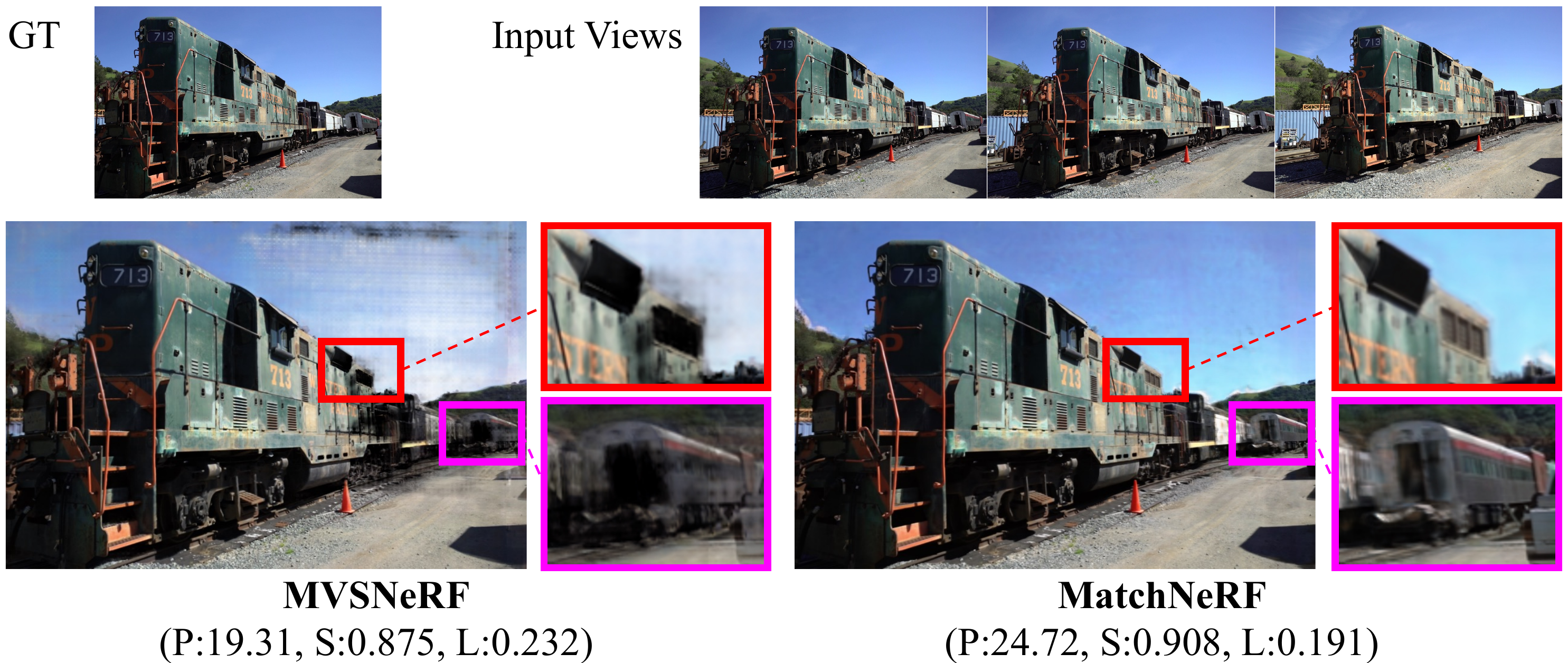}
    \end{center}
    \vspace{-1em}
    \caption{\red{\textbf{Qualitative results on Tanks and Temples}. Views are from the ``Train'' scene. Input views include the 3 viewpoints nearest to the target view, with the first input view serving as the reference view for MVSNeRF. Quantitative results are placed below each image, showing the scores for PSNR, SSIM, and LPIPS. MatchNeRF renders better details with a cleaner background compared to MVSNeRF.} 
    }
    \label{fig:tnt_comp}
\end{figure}

\red{
\subsubsection{Comparisons on More Challenging Dataset}
To further confirm that our MatchNeRF can perform well on more complex and larger-scale scenes, we conduct experiments on the Tanks and Temples (T\&T)~\cite{knapitsch2017tanks} dataset.
We follow a similar evaluation setup as in the RFF~\cite{mildenhall2020nerf} dataset. In particular, for each scene, 4 randomly selected target views are rendered, with an image resolution of $640 \times 960$.
The 3 viewpoints nearest to the target view are selected as input.
Note that we used the same model trained on DTU as in the previous experiments and tested it directly on T\&T without any fine-tuning.

As reported in TABLE~\ref{tab:tnt_comp}, our MatchNeRF significantly outperforms the main comparison method, MVSNeRF~\cite{chen2021mvsnerf}, suggesting that the introduced explicit correspondence matching is more robust in handling challenging cases by effectively fusing information from all the selected nearby input views.
We observe that MatchNeRF reconstructs details and background with reasonably good quality (see Fig.\ref{fig:tnt_comp}), even when camera poses may be slightly inaccurate, as T\&T only provides pseudo ground truth camera poses obtained via COLMAP. The main reason is that our correspondence matching operates in the feature space, where larger receptive fields make the model more robust to slightly noisy data.}

\minhl{
To provide a more comprehensive comparison, we also report efficiency in terms of training and testing runtimes.
Specifically, we measure the average training iteration time using the default setting (512 rays per batch) and the average inference time over a test batch (4096 rays, consistent with TABLE~\ref{tab:moreviews}).
As reported in TABLE~\ref{tab:tnt_comp}, our MatchNeRF runs slightly slower than MVSNeRF during both training and testing.
This is expected, as our method builds on a Transformer-based architecture, whereas MVSNeRF relies on a CNN backbone.
The efficiency gap may be reduced by incorporating more optimized attention implementations~\cite{dao2023flashattention2}.
Despite this trade-off, MatchNeRF offers clear advantages in visual and geometric quality, as well as generalization capability, as demonstrated in all the above analyses.
}

\begin{table}[t]
\caption{\textbf{Quantitative comparison on Tanks and Temples}~\cite{knapitsch2017tanks}. \red{MatchNeRF outperforms MVSNeRF on the challenging T\&T dataset, highlighting the effectiveness of the correspondence matching. Both models are trained on DTU~\cite{jensen2014large} and directly tested on T\&T. We select the 3 viewpoints nearest to the target view as input and evaluate only on the central regions. \minhl{We also report efficiency comparison regarding training runtime per iteration (\emph{512-rays}) and inference runtime per test batch (\emph{4096-rays}).}} 
\label{tab:tnt_comp}}
\begin{center}
\resizebox{\linewidth}{!}{%
\begin{tabular}{lccccc} 
\toprule
Method  & PSNR$\uparrow$ & SSIM$\uparrow$ & LPIPS$\downarrow$  & \minhl{Train $t\downarrow$} & \minhl{Test $t\downarrow$}  \\ 
\midrule
MVSNeRF~\cite{chen2021mvsnerf}       & 19.77          & 0.827          & 0.272     & \minhl{\textbf{0.113 s}}  & \minhl{\textbf{0.056 s}}      \\
MatchNeRF     & \textbf{21.94}          & \textbf{0.840}          & \textbf{0.258}   & \minhl{0.145 s} &   \minhl{0.081 s}  \\
\bottomrule
\end{tabular}
}
\end{center}
\end{table}

\subsection{Per-scene Fine-tuning Experiment \label{sec:app_finetune}}

Although MatchNeRF is designed for generalizing to new, unseen scenes, it can also incorporate an additional per-scene fine-tuning stage.
We follow the settings of MVSNeRF to fine-tune our model on each specific DTU test scene with 16 additional training views.
As reported in TABLE~\ref{tab:finetune}, MatchNeRF again outperforms its main competitor MVSNeRF under the same 10k iterations of per-scene fine-tuning, with a shorter optimization time (16-mins vs. 24-mins).
Note that despite outstanding per-scene fine-tuning performance,  Point-NeRF~\cite{xu2022point} ($23.89/0.874/0.203$) performs much worse than MVSNeRF($26.63/0.931/0.168$) and our MatchNeRF ($26.91/0.934/0.159$) under our main focus of \emph{generalizable setting}.

\begin{table}[t]
\caption{\textbf{Per-scene Fine-tuning Comparisons on DTU}. Despite outstanding results with fine-tuning, Point-NeRF is much worse than our MatchNeRF under our main focus of \emph{generalizable setting}. The results of other models are all borrowed from Point-NeRF's paper, measured over \emph{the foreground region}. \label{tab:finetune}}
\vspace{-15pt}
\begin{center}
\resizebox{\linewidth}{!}
{%
\begin{tabular}{lcccl}
\toprule
Method\textsubscript{iterations}   & PSNR$\uparrow$ & SSIM$\uparrow$ & LPIPS$\downarrow$ & \multicolumn{1}{c}{Time$\downarrow$}     \\
\midrule
NeRF\textsubscript{200k} & 27.01       & 0.902   & 0.263   & 10 hr  \\
Point-NeRF\textsubscript{10k} & 30.12  & 0.957  & 0.117   & 20 min \\
MVSNeRF\textsubscript{10k} & 28.50     & 0.933  & 0.179   & 24 min \\
MatchNeRF\textsubscript{10k}  & 28.53       & 0.938  & 0.170   & 16 min \\
\bottomrule
\end{tabular}
}
\end{center}
\end{table}

\section{Conclusion}
\label{sec:conclusion}
We have proposed a new generalizable NeRF method that uses explicit correspondence matching statistics as the geometry prior.
We implemented the matching process by applying group-wise cosine similarity over pair-wise features that are enhanced with a cross-view Transformer.
Unlike the prior cost-volume-based approaches that are inherently limited by the selection of the reference view, our matching-based design is view-agnostic.
We showcased that our explicit correspondence matching information can provide valuable geometry cues for the estimation of volume density, leading to state-of-the-art performance under several challenging settings on three benchmark datasets. 

Although we have demonstrated the superior performance of our MatchNeRF under various challenging yet practical \emph{few-view} (\eg, 2 or 3) scenarios, 
we acknowledge that our current architecture might be less effective at handling occlusions since they have not been explicitly modeled. 
These occlusion issues may become severe when the input contains a large number of views (\eg, $\geq 10$).
We believe that MatchNeRF can serve as a novel basis for further related research, \eg, enhancing it by explicitly modeling the occlusion factor.
In addition, combining our explicit correspondence matching design with fast optimization NeRF approaches or sparse views settings will also be a promising direction. \red{Extending MatchNeRF to handle noisy or even missing camera poses would make it more practical.} \minhl{Furthermore, the core concept of using feature matching as a geometry prior may continue to benefit feed-forward 3D reconstruction as more advanced 3D representations emerge~\cite{chen2024mvsplat}.}

\appendices

\section{Limitation of Cost Volume \label{sec:app_cv_limit}}
\begin{figure}[ht]
    \begin{center}
    \includegraphics[width=\linewidth]{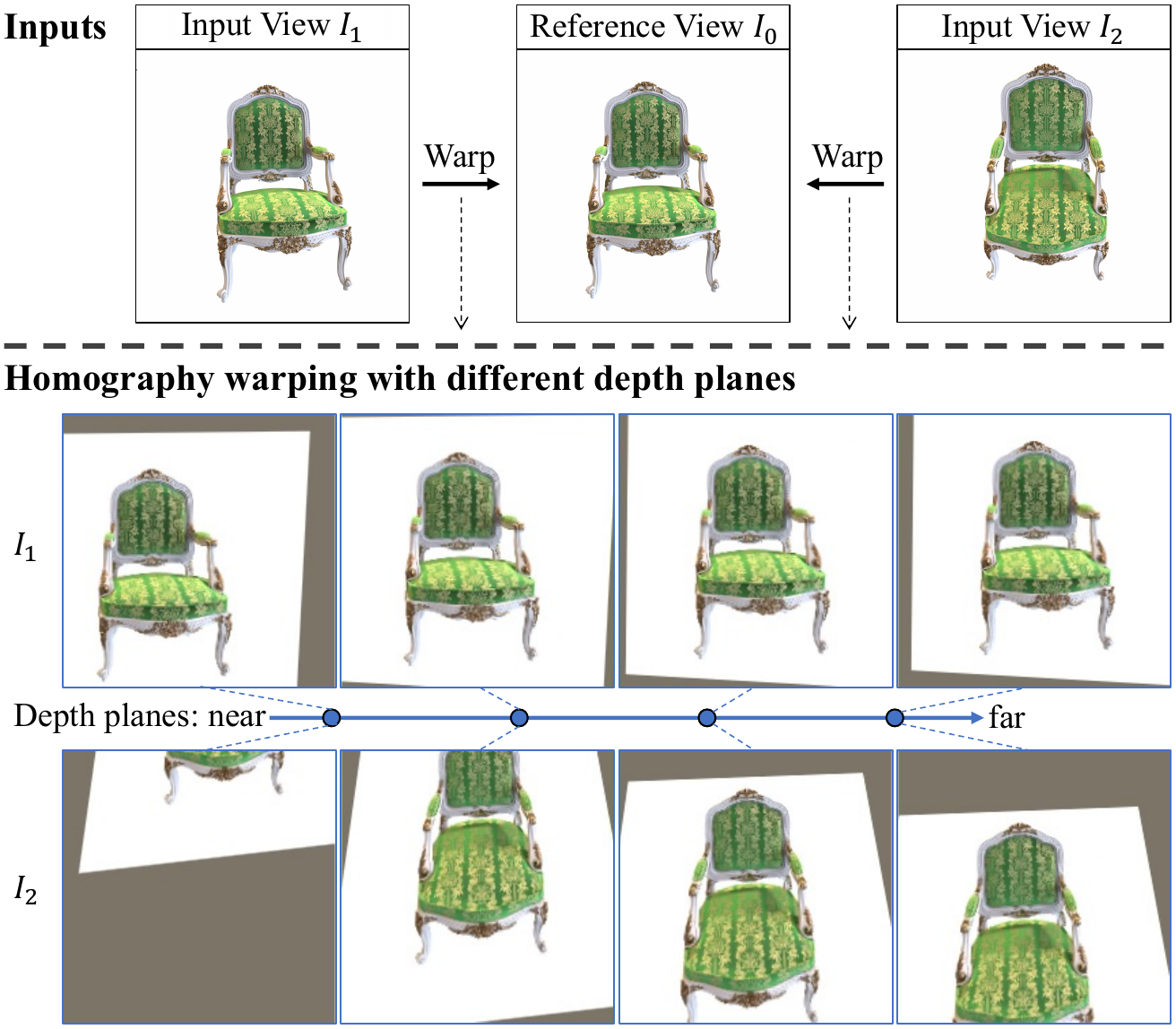}
    \end{center}
    \caption{\textbf{Illustration of warping two input views to the reference view with different depth planes}. The construction of cost volume relies on warping different input views to the reference view with different depth planes, where inevitable artifacts will be introduced in the background due to the warping operation.  
    }
    \label{fig:warp_imgs}
\end{figure}

To better understand the limitation of cost volume, we further visualize the warped input views in Fig.~\ref{fig:warp_imgs}. The construction of cost volume relies on warping different input views to one selected reference view with different depth planes, where artifacts will be inevitably introduced in the background of those warped views. Thus, the quality of the cost volume is inherently limited, accordingly leading to artifacts in the background of the rendered image (see Fig.~\ref{tab:sota}). In contrast, our method avoids the warping and cost volume construction operations, thus achieving better performance both qualitatively and quantitatively.

\section{Backbone Training Strategy \label{sec:app_backbone}}
As stated in Sec.~\ref{sec:exp_implement}, our feature extractor is initialized with GMFlow's released weights. We observe that such a \emph{publicly available} pretrained model can serve as \emph{a good initialization} to our encoder, hence saving us time in training the full model.
We further compare with other training strategies, so as to better demonstrate the effectiveness of our default model (TABLE~\ref{tab:bb_init} `pretrained + finetune').

\textbf{Is GMFlow's released weights effective enough to be directly used (without any finetuning) in our task? NO.} 
MatchNeRF performs poorly if we \emph{freeze} the backbone initialized with the pretrained weights and train solely the NeRF decoder (TABLE~\ref{tab:bb_init} `pretrained + freeze'). The main reason is that GMFlow is trained on Sintel for optical flow estimation, whose full model and final objective share little in common with ours. In particular, Sintel has \emph{no overlap} with any data we used, and its contents are notably different from ours (video game scenes vs. daily life scenes). Moreover, MatchNeRF is built for novel view synthesis rather than optical flow estimation, the differences between these two tasks further reduce the effectiveness of GMFlow's pretrained model. 
The key observation is that our \emph{backbone features} should be aligned across different views, which does share similar objective with that of GMFlow's \emph{backbone}, thus its pretrained model can serve as a good initialization but requires further fine-tuning to adapt to our task.

\textbf{Can our MatchNeRF be trained from scratch? YES.} 
We retrain our MatchNeRF by randomly initializing the whole framework. The ablation model (TABLE~\ref{tab:bb_init} `random init + finetune') achieves comparable results with our default model, which again outperforms our main competitor MVSNeRF ($20.67/0.865/0.296$). It demonstrates that the \emph{cross-view interactions} can be learned from scratch 
without any explicit correspondence supervision. 

In general, our MatchNeRF can be trained from scratch to achieve state-of-the-art performance, and it can be further enhanced by leveraging the publicly available GMFlow pretrained model as an initialization.

\begin{table}[t]
\caption{\textbf{Backbone Training Strategy}. The results are reported on DTU using 3 nearest input views, measured over \emph{the whole image}. The settings `random init' and `pretrained' denote the backbone is `randomly initialized' and `initialized with GMFlow pretrained weights', respectively. The settings `freeze' and `finetune' refer to `freeze' and `finetune' the trainable parameters of the backbone, respectively. Although MatchNeRF can be trained from scratch to reach state-of-the-art performance, leveraging the publicly available pretrained weights as an initialization leads to better results.
\label{tab:bb_init}}
\begin{center}
\resizebox{0.9\linewidth}{!}{%
\begin{tabular}{lccc} 
\toprule
Setting  & PSNR$\uparrow$ & SSIM$\uparrow$ & LPIPS$\downarrow$  \\ 
\midrule
pretrained + freeze & 20.36          & 0.845          & 0.318              \\
random init + finetune     & 23.44          & 0.881          & 0.283              \\
pretrained + finetune    & \textbf{24.54} & \textbf{0.897} & \textbf{0.257}     \\
\bottomrule
\end{tabular}}
\end{center}

\end{table}

\section*{Acknowledgments}
This research is supported by the Monash FIT Start-up Grant. Dr. Chuanxia Zheng is supported by EPSRC SYN3D EP/Z001811/1.

\ifCLASSOPTIONcaptionsoff
  \newpage
\fi

{
    \bibliographystyle{IEEEtran}
    \bibliography{main}
}

\end{document}